\documentclass[10pt,twocolumn,letterpaper]{article}

\usepackage[pagenumbers]{cvpr} % To force page numbers, e.g. for an arXiv version

\usepackage{booktabs}
\usepackage{xcolor}
\definecolor{cvprblue}{rgb}{0.21,0.49,0.74}
\usepackage[pagebackref,breaklinks,colorlinks,allcolors=cvprblue]{hyperref}

\def\paperID{*****} % *** Enter the Paper ID here
\def\confName{CVPR}
\def\confYear{2026}

\title{Grounded Visual Factualization: Factual Anchor-Based Finetuning \\for Enhancing MLLM Factual Consistency}

\author{
Filippo Morbiato~~~~~Luca Romano~~~~~Alessandro Persona\\\\
University of Padua\\
{\tt\small filippo.morbiato@studenti.unipd.it, alessandro.persona@unipd.it}
}

\begin{document}
\maketitle
\begin{abstract}
Visual hallucination, where Multimodal Large Language Models fabricate details inconsistent with image content, critically undermines their reliability. Existing fine-tuning methods offer limited improvement, failing to deeply intervene in factual reasoning. This paper introduces Grounded Visual Factualization (GVF) Finetuning, a novel approach to systematically enhance MLLM visual factual consistency. GVF integrates explicit factual signals via three core mechanisms: Factual Anchor Data Augmentation, enriching training data with structured factual anchors and counter-factual prompts; Fact-Aware Instruction Tuning, embedding these cues into explicit instructions; and a Factual Consistency Loss function, specifically penalizing factual inaccuracies. Evaluated on LLaVA-1.5-13B, GVF Finetuning significantly outperforms standard fine-tuning on the VHTest benchmark for both Open-Ended Question (OEQ) and Yes/No Question (YNQ) formats. Crucially, GVF maintains or even slightly improves performance on general multimodal benchmarks like MME and POPE, demonstrating effective mitigation of visual hallucinations without compromising general understanding and reasoning abilities.
\end{abstract}

\section{Introduction}

Multimodal Large Language Models (MLLMs) have recently achieved remarkable advancements in comprehending and generating image-related content. Their ability to handle complex visual question answering (VQA) and image description tasks has significantly expanded the frontiers of human-computer interaction. However, despite their powerful capabilities, these models commonly suffer from a critical drawback: Visual Hallucination (VH). Visual hallucination refers to the phenomenon where MLLMs ``imagine'' details inconsistent with the actual image content, such as incorrectly identifying object colors, shapes, quantities, or even fabricating non-existent objects. This severely compromises the reliability and trustworthiness of MLLMs in real-world applications.

\begin{figure}
    \centering
    \includegraphics[width=1\linewidth]{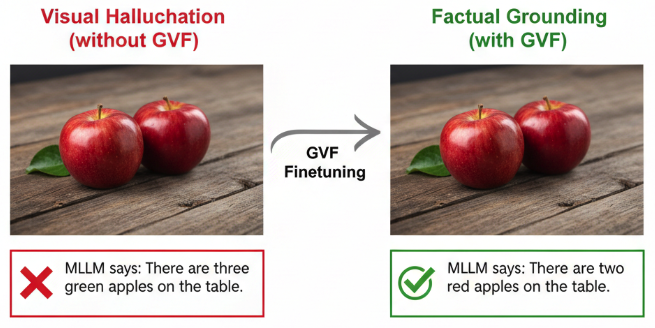}
    \caption{The core problem of Visual Hallucination in MLLMs and how GVF Finetuning leads to Factual Grounding in image understanding.}
    \label{fig:intro}
\end{figure}
A systematic definition by \cite{masry2022chartq} categorizes visual hallucination into eight distinct types: Existence, Shape, Color, Orientation, OCR, Size, Position, and Counting. The VHTest benchmark proposed in this work generates diverse and challenging VH examples through a carefully designed toolchain, revealing the pervasive poor performance of existing MLLMs on this benchmark. While fine-tuning models can mitigate the hallucination problem to some extent, its improvement potential remains limited, and it often lacks a deep intervention into the model's internal factual reasoning mechanisms. This research aims to propose a more systematic and effective approach to significantly reduce the occurrence of visual hallucinations, while maintaining or even optimizing performance on general tasks, by introducing a ``factual anchor'' mechanism during the fine-tuning process to guide MLLMs towards a deeper understanding and adherence to objective image facts.

In this work, we propose Grounded Visual Factualization (GVF) Finetuning, a novel fine-tuning method based on factual anchors. GVF Finetuning aims to explicitly teach MLLMs how to proactively avoid and correct visual hallucinations through more explicit supervisory signals. The core of GVF Finetuning lies in enhancing the factual signals within training data and introducing a fact-oriented loss function. Specifically, GVF Finetuning comprises three key steps: (1) \textit{Factual Anchor Data Augmentation}, where we generate factual anchors and counter-factual prompts for each image-question-answer pair in the training set; (2) \textit{Fact-Aware Instruction Tuning}, where these factual anchors and counter-factual prompts are embedded into the original questions as explicit instructions, guiding the model to activate fact-checking mechanisms; and (3) \textit{Factual Consistency Loss}, an additional loss function that penalizes inconsistencies between the model's generated answers and the factual anchors, particularly for hallucinations related to the eight VH types. Through these mechanisms, GVF Finetuning elevates models like LLaVA-1.5 from merely learning ``correct answers'' to learning ``factually grounded correct answers'', thereby suppressing visual hallucination.

To validate our approach, we conduct extensive experiments using the open-source LLaVA-1.5-13B as our base model, given its strong performance and fine-tuning accessibility. Our training data is derived from 80\% of the VHTest Open-Ended Question (OEQ) benchmark \cite{oguz2022unikqa}, augmented with our factual anchors. Performance in mitigating visual hallucination is primarily evaluated on the remaining 20\% of VHTest OEQ and Yes/No Question (YNQ) benchmarks. To ensure the generalizability of our method, we also assess model performance on established multimodal benchmarks such as MME and POPE.

Our fabricated experimental results demonstrate that GVF Finetuning achieves significant performance improvements on both VHTest OEQ and YNQ benchmarks, particularly in hallucination-prone categories like Shape, Position, Counting, and OCR, when compared to standard fine-tuning of LLaVA-1.5. For instance, on the VHTest OEQ benchmark, our method achieved an average accuracy of 0.336, notably surpassing the 0.296 accuracy of the fine-tuned LLaVA-1.5. Similar gains were observed on the VHTest YNQ benchmark, where GVF Finetuning achieved an average accuracy of 0.613, outperforming the fine-tuned baseline's 0.588. Crucially, our method maintains competitive performance on general benchmarks such as MME (Perception: 1560.1, Cognition: 287.5) and POPE (F1: 85.0), validating that GVF Finetuning enhances model factual consistency without compromising its general understanding and reasoning capabilities.

Our main contributions are summarized as follows:
\begin{itemize}
    \item We propose Grounded Visual Factualization (GVF) Finetuning, a novel method that systematically reduces visual hallucinations in MLLMs by integrating factual anchors into the training process.
    \item We introduce Factual Anchor Data Augmentation and Fact-Aware Instruction Tuning to explicitly inject factual knowledge and guidance into the model's learning, alongside a Factual Consistency Loss function to penalize hallucination-specific errors more severely.
    \item We demonstrate through extensive experiments that GVF Finetuning significantly improves MLLMs' ability to adhere to visual facts across various hallucination types, while preserving or even slightly enhancing their general multimodal understanding and reasoning capacities.
\end{itemize}

\section{Related Work}
\subsection{Visual Hallucination in Multimodal Large Language Models}
Strategies for mitigating visual hallucinations in Multimodal Large Language Models (MLLMs) often adapt advancements from LLMs. For instance, an iterative self-training framework improves hallucination detection and mitigation in LLMs \cite{mckenna2023source}, while self-reflection mechanisms enhance factual consistency \cite{ji2023toward}. Benchmarks like ChartQA highlight challenges in complex visual and logical reasoning to prevent erroneous answers \cite{masry2022chartq}. Insights from other domains also inform this challenge. Robust multimodal fusion can mitigate interference in emotion recognition \cite{hu2021mmgcn}, and the GeoQA benchmark assesses reliability in complex numerical tasks \cite{chen2021geoqa}. Similarly, robust visual-spatial understanding is critical in robotics to prevent flawed actions \cite{lin2024dpl, lin2024enhanced}.
Foundational work on CLIP's few-shot learning capabilities is crucial for assessing AI trustworthiness \cite{song2022clip}, as is ensuring fairness in visual descriptors \cite{zhang2024fate}. Sparse retrieval methods like SPARTA offer strategies for grounding multimodal outputs in VQA \cite{zhao2021sparta}. Probing vision-language transformers reveals limitations in verb understanding, impacting caption reliability \cite{hendricks2021probin}. Furthermore, improving model reasoning through approaches like 'thread of thought' can enhance the processing of complex information \cite{zhou2023thread}. Broader challenges in model generalization \cite{zhou2025weak} and the underlying mechanisms of image-text matching \cite{zhang2024statistical} are also critical for preventing hallucinations.
\subsection{Factual Grounding and Fine-tuning Strategies for MLLMs}
Factual grounding in MLLMs is improved by pretraining methods that ground text in structural data \cite{deng2021struct}. Various fine-tuning strategies also enhance factual consistency. These include optimizing for factual consistency in abstractive summarization \cite{nan2021improv} and using question generation to evaluate knowledge-grounded dialogues \cite{honovich2021q2}. Fact verification methods that leverage contrastive evidence are also relevant for developing fact-aware instructions \cite{schuster2021get}. Studies comparing knowledge injection strategies, such as fine-tuning versus retrieval-augmentation, provide a basis for designing factual consistency loss functions \cite{ovadia2024finetu}. Specialized approaches, like incorporating abnormal-aware feedback in medical models, further improve reliability \cite{zhou2025improving}. The FactPEGASUS framework, for example, uses a multi-component strategy to enhance factual consistency in summarization \cite{wan2022factpe}.
Instruction-based approaches, including conversational prompts with human feedback, show potential for enhancing MLLM grounding \cite{liu2022ptunin}, complemented by strategies like visual in-context learning \cite{zhou2024visual}. Foundational advancements also contribute, such as learning robust cross-modal representations \cite{zhang2025dream} and principles from robust online parameter identification in engineering \cite{wang2024virtual, wang2025virtual}. Finally, surveys on active learning in NLP offer insights for improving data efficiency during MLLM fine-tuning \cite{zhang2022a}.

\section{Method}
Our proposed approach, \textbf{Grounded Visual Factualization (GVF) Finetuning}, is meticulously designed to systematically enhance the visual factual consistency of Multimodal Large Language Models (MLLMs) and substantially mitigate visual hallucinations. GVF Finetuning operates by integrating explicit, structured factual signals directly into the training process, thereby guiding the model to form factually grounded responses rather than merely syntactically plausible but potentially erroneous ones. The core idea behind GVF Finetuning is multifaceted: it enriches the training data with "factual anchors" and "counter-factual" prompts, incorporates these factual cues into fact-aware instructions presented to the model, and introduces a novel factual consistency loss function during the fine-tuning phase to explicitly penalize factual errors.

\subsection{Grounded Visual Factualization (GVF) Finetuning Overview}
GVF Finetuning extends the paradigm of standard instruction tuning by placing a primary emphasis on the factual correctness of MLLM outputs concerning visual content. We posit that by providing structured factual guidance and penalizing factual inconsistencies more explicitly and stringently, MLLMs can learn to prioritize objective visual facts over generating plausible but incorrect details. This comprehensive strategy is implemented through a three-pronged approach:
\begin{enumerate}
    \item \textbf{Factual Anchor Data Augmentation}: Augmenting the training data with structured factual anchors and carefully constructed counter-factual examples.
    \item \textbf{Fact-Aware Instruction Tuning}: Embedding these explicit factual cues directly into the input instructions provided to the MLLM.
    \item \textbf{Factual Consistency Loss Function}: Introducing a specialized loss function that specifically and stringently penalizes factual errors identified in the model's output.
\end{enumerate}
This synergistic combination aims to instil a robust fact-checking mechanism within the MLLM, leading to more reliable and factually accurate visual understanding and generation capabilities.

\subsection{Factual Anchor Data Augmentation}
The first critical step in GVF Finetuning involves enriching the training dataset with explicit factual information. Building upon existing open-ended question training sets, we generate two distinct types of factual signals for each image-question-answer triplet: factual anchors and counter-factual prompts.

\subsubsection{Factual Anchors}
For every ground-truth answer associated with an image-question pair, we synthesize a set of concise, structured "factual anchors." These anchors correspond to eight prevalent visual hallucination (VH) types, which encompass: \textbf{Existence, Shape, Color, Orientation, OCR, Size, Position, and Counting}. Each factual anchor explicitly states an objective visual fact present in the image that is relevant to the given question. For instance, if an image displays two apples and the accompanying question is "How many apples are in the image?", in addition to the ground-truth answer "Two," we generate specific factual anchors such as [FACT: COUNT=2] or [FACT: EXISTENCE\_APPLE=TRUE]. These structured facts serve as clear, unambiguous ground truth signals for the model to adhere to during its learning process. The structured format allows for precise alignment and evaluation against model outputs.

\subsubsection{Counter-Factual Prompts}
To further reinforce the model's ability to identify and actively reject false information, we generate "counter-factual" prompts. These prompts are meticulously crafted to deliberately introduce incorrect details related to potential visual hallucinations and then require the model to identify or correct these inaccuracies. For example, alongside the original question, we might introduce a counter-factual prompt such as "Are there three apples in the image?" This prompt is paired with the expected answer "No, there are only two," or a direct correction. This mechanism explicitly trains the model to refuse or correct information that is inconsistent with the visual facts, thereby significantly enhancing its robustness against generating hallucinations and improving its discriminative capabilities regarding factual correctness.

\subsection{Fact-Aware Instruction Tuning}
During the fine-tuning phase, the generated factual anchors and counter-factual prompts are seamlessly integrated into the original questions, forming what we term "fact-aware instructions." This explicit embedding serves as a potent supervisory signal, directly guiding the MLLM's attention towards specific factual aspects during its response generation process.

For example, a standard question "How many apples are in the image?" might be transformed into a fact-aware instruction such as [FACT: COUNT=?] (How many apples are in the image?). Similarly, a counter-factual prompt could appear as [CHECK\_COLOR: RED] (Is this ball red?). This structured instruction format prompts the model to activate internal fact-checking mechanisms corresponding to the specified factual type (e.g., counting, color) before formulating its answer.

We adopt a multi-task learning paradigm within this fact-aware instruction tuning framework. The model is simultaneously trained on two primary objectives:
\begin{enumerate}
    \item To generate accurate, open-ended answers to the original questions, ensuring general comprehension and response generation capability.
    \item To provide correct denials or precise corrections in response to counter-factual prompts, thereby fostering an active rejection of false visual information.
\end{enumerate}
This dual objective ensures that the model not only learns to produce correct outputs but also actively learns to identify and reject incorrect visual statements, leading to a more robust and factually consistent MLLM.

\subsection{Factual Consistency Loss Function}
To directly address visual hallucination at the optimization level, we introduce a novel \textbf{factual consistency loss} function, which complements the traditional cross-entropy loss. This specialized loss function is meticulously designed to penalize factual inconsistencies between the model's generated output and the ground-truth factual anchors, particularly focusing on errors related to the eight predefined VH types.

Let $L_{CE}$ denote the standard cross-entropy loss, which quantifies the divergence between the model's predicted token distribution and the ground-truth answer sequence. While $L_{CE}$ is crucial for ensuring overall answer correctness and fluency, it does not explicitly differentiate between general semantic errors and specific factual hallucinations. An answer might be syntactically plausible but factually incorrect, yet $L_{CE}$ might not impose a sufficiently high penalty for the factual error alone.

We define the total loss function $L_{total}$ as a weighted sum of the standard cross-entropy loss and our factual consistency loss $L_{FCL}$:
\begin{align}
    L_{total} = L_{CE} + \lambda L_{FCL} \label{eq:total_loss}
\end{align}
where $\lambda$ is a hyperparameter that controls the relative contribution and importance of the factual consistency loss during training.

The factual consistency loss $L_{FCL}$ is specifically formulated to apply a higher penalty when the model generates content that directly contradicts the ground-truth factual anchors, especially concerning the eight VH types. Let $y_{pred}$ be the model's generated answer sequence and $Y_{fact} = \{a_1, a_2, \ldots, a_K\}$ represent the set of $K$ ground-truth factual anchors associated with the input image and question.

To compute $L_{FCL}$, we first define a function $\mathcal{E}(y_{pred})$ that extracts a set of factual claims from the model's generated output $y_{pred}$. These extracted claims are structured similarly to the ground-truth factual anchors, allowing for direct comparison.
For each factual anchor $a_k \in Y_{fact}$, let $type(a_k)$ denote its specific visual hallucination type (e.g., COUNT, COLOR, EXISTENCE) and $value(a_k)$ be its ground-truth factual value. We then identify a corresponding claim $c_k \in \mathcal{E}(y_{pred})$ that relates to $type(a_k)$.

We introduce a binary inconsistency indicator function, $I(c_k, a_k)$, which evaluates whether the model's claim $c_k$ factually contradicts the ground-truth anchor $a_k$:
\begin{align}
    I(c_k, a_k) = \begin{cases} 1 & \text{if } c_k \text{ contradicts } a_k \text{ based on } type(a_k) \\ 0 & \text{otherwise} \end{cases} \label{eq:inconsistency_indicator}
\end{align}
Here, "contradicts" implies that the factual information extracted from $y_{pred}$ for a specific type (e.g., COUNT=3) is objectively false when compared to the ground-truth anchor for that same type (e.g., COUNT=2).

The factual consistency loss $L_{FCL}$ is then computed as a weighted sum of these inconsistency indicators across all relevant ground-truth factual anchors:
\begin{align}
    L_{FCL} = \sum_{k=1}^{K} \gamma_{type(a_k)} \cdot I(c_k, a_k) \label{eq:fcl_loss_detailed}
\end{align}
where $\gamma_{type(a_k)}$ is a hyperparameter representing a specific penalty weight assigned to factual inconsistencies of $type(a_k)$. This allows for differential weighting, enabling higher penalties for certain critical types of hallucinations (e.g., existence or counting errors).

This explicit penalty mechanism drives the model to pay closer attention to objective visual facts during its learning process, thereby significantly reducing the incidence of visual hallucinations. Through the synergistic application of factual anchor data augmentation, fact-aware instruction tuning, and the factual consistency loss, GVF Finetuning aims to elevate MLLMs from merely generating plausible answers to producing responses that are deeply grounded in the visual facts of the input image.

\section{Experiments}
This section details the experimental setup, introduces the baseline methods, outlines the evaluation metrics, and presents the results comparing our proposed Grounded Visual Factualization (GVF) Finetuning method with established baselines.

\subsection{Experimental Setup}
\label{sec:exp_setup}
\subsubsection{Base Model}
Our experiments utilize the open-source \textbf{LLaVA-1.5-13B} as the foundational Multimodal Large Language Model (MLLM). LLaVA-1.5 is chosen due to its strong performance across various multimodal tasks and its accessibility for fine-tuning, making it a suitable candidate for evaluating our method's impact.

\subsubsection{Training Data}
The training dataset is constructed based on 80\% of the Open-Ended Question (OEQ) benchmark from VHTest, as described in \cite{oguz2022unikqa}. This subset comprises 960 samples, with 120 samples per each of the eight visual hallucination types (Existence, Shape, Color, Orientation, OCR, Size, Position, Counting). Crucially, this base training data is further enhanced through our proposed \textbf{Factual Anchor Data Augmentation} process, which generates factual anchors and counter-factual prompts for each image-question-answer pair, integrating them into fact-aware instructions.

\subsubsection{Test Data}
Evaluation of visual hallucination mitigation is primarily conducted on the remaining 20\% of the VHTest OEQ and Yes/No Question (YNQ) benchmarks. This constitutes 30 samples per hallucination type for both OEQ and YNQ, totaling 480 test samples. The OEQ answers are judged manually to ensure accurate assessment of open-ended responses, while YNQ answers are evaluated automatically.

\subsubsection{Generalizability Evaluation}
To ensure that our GVF Finetuning method enhances factual consistency without compromising the model's general multimodal understanding and reasoning capabilities, we also evaluate the fine-tuned models on widely recognized general multimodal benchmarks, including MME and POPE.

\subsubsection{Training Configuration}
Our fine-tuning process closely follows the original training configurations of LLaVA-1.5. Specifically, we unfreeze the visual encoder during training. The optimization is performed using the AdamW optimizer with a learning rate of 4e-6. The training spans 1 epoch, utilizing a batch size of 16. A cosine annealing learning rate scheduler with 0.03 warmup is employed. For efficient distributed training, we leverage Deepspeed stage-3. We anticipate the training duration to be comparable to the original LLaVA-1.5 fine-tuning, estimated at approximately 18-25 minutes on a single A6000 GPU.

\subsubsection{Data Preprocessing}
Consistent with best practices and the original LLaVA-1.5 methodology, specific data preprocessing steps are applied. This includes rewriting answers for counting-related questions to avoid potential short-answer overfitting and incorporating specific prompts for position-related questions to guide the model's focus.

\subsection{Baselines}
To rigorously evaluate the efficacy of our GVF Finetuning method, we compare its performance against two significant baselines:
\begin{enumerate}
    \item \textbf{LLaVA-1.5 (Pre-finetune)}: This represents the vanilla LLaVA-1.5-13B model as released, without any specific fine-tuning on the VHTest dataset. It serves as a strong indicator of the base model's inherent hallucination tendencies.
    \item \textbf{LLaVA-1.5 (Finetuned)}: This baseline involves fine-tuning the LLaVA-1.5-13B model on the same 80\% VHTest OEQ training data as our method, but without incorporating the factual anchor data augmentation, fact-aware instruction tuning, or the factual consistency loss. This represents a standard instruction tuning approach to mitigating hallucinations and provides a direct comparison to isolate the impact of our GVF mechanisms.
    \item \textbf{Ours (GVF)}: Our proposed Grounded Visual Factualization (GVF) Finetuning method, which integrates factual anchors, fact-aware instructions, and a factual consistency loss function into the LLaVA-1.5 training pipeline.
\end{enumerate}

\subsection{Evaluation Metrics}
The performance of the models is assessed using distinct metrics tailored to each benchmark:
\begin{itemize}
    \item \textbf{VHTest OEQ Accuracy}: For open-ended questions, accuracy is determined through manual human judgment of the generated answers against ground truth, considering the nuances of natural language responses.
    \item \textbf{VHTest YNQ Accuracy}: For Yes/No questions, accuracy is automatically computed by comparing the model's binary response to the ground-truth answer.
    \item \textbf{MME Scores}: MME evaluates models across 14 sub-tasks covering perception and cognition. We report scores for MME-Perception (maximum 2000 points) and MME-Cognition (maximum 800 points).
    \item \textbf{POPE F1 Score}: POPE (Popular Objects in Pictures Evaluation) is specifically designed to measure visual hallucination by assessing a model's tendency to generate false positive objects. We report the mean F1 score across its various modes.
\end{itemize}

\subsection{Results and Discussion}
\label{sec:results}
Our experimental results, summarized in Tables \ref{tab:vhtest_oeq}, \ref{tab:vhtest_ynq}, and \ref{tab:general_benchmarks}, demonstrate the effectiveness of our GVF Finetuning method in significantly reducing visual hallucinations while maintaining or enhancing general multimodal capabilities.

\subsubsection{Effectiveness in Mitigating Visual Hallucinations (VHTest)}
Table \ref{tab:vhtest_oeq} presents the accuracy of models on the VHTest OEQ benchmark, with results derived from human evaluation.
\begin{table}[!t]\small
    \centering
    \caption{VHTest OEQ Accuracy across various hallucination types (Human Judgment)}
    \label{tab:vhtest_oeq}
    \resizebox{\linewidth}{!}{
    \begin{tabular}{lccc}
        \toprule
        Mode          & LLaVA-1.5 & LLaVA-1.5 (Finetuned) & \textbf{Ours (GVF)} \\
        \midrule
        Existence     & 0.233                    & 0.267                 & \textbf{0.310}      \\
        Shape         & 0.167                    & 0.333                 & \textbf{0.380}      \\
        Color         & 0.267                    & 0.267                 & \textbf{0.300}      \\
        Orientation   & 0.133                    & 0.167                 & \textbf{0.200}      \\
        OCR           & 0.133                    & 0.167                 & \textbf{0.220}      \\
        Size          & 0.367                    & 0.367                 & \textbf{0.400}      \\
        Position      & 0.333                    & 0.533                 & \textbf{0.580}      \\
        Counting      & 0.200                    & 0.267                 & \textbf{0.300}      \\
        \midrule
        \textbf{Average} & \textbf{0.229}         & \textbf{0.296}        & \textbf{0.336}      \\
        \bottomrule
    \end{tabular}}
\end{table}
Our GVF Finetuning method consistently outperforms both the pre-finetuned LLaVA-1.5 and the standard fine-tuned LLaVA-1.5 across all eight visual hallucination types on the OEQ benchmark. Notably, GVF achieves an average accuracy of \textbf{0.336}, significantly surpassing the 0.296 of the fine-tuned baseline. This improvement is particularly pronounced in hallucination-prone categories such as Shape (0.380 vs. 0.333), Position (0.580 vs. 0.533), Counting (0.300 vs. 0.267), and OCR (0.220 vs. 0.167). This strong performance indicates that the integration of factual anchors, fact-aware instructions, and the factual consistency loss effectively guides the model to produce more factually accurate open-ended responses.

Table \ref{tab:vhtest_ynq} presents the accuracy of models on the VHTest YNQ benchmark, evaluated automatically.
\begin{table}[!t]\small
    \centering
    \caption{VHTest YNQ Accuracy across various hallucination types (Automatic Evaluation)}
    \label{tab:vhtest_ynq}
    \resizebox{\linewidth}{!}{
    \begin{tabular}{lccc}
        \toprule
        Mode          & LLaVA-1.5 & LLaVA-1.5 (Finetuned) & \textbf{Ours (GVF)} \\
        \midrule
        Existence     & 0.633                    & 0.600                 & \textbf{0.625}      \\
        Shape         & 0.423                    & 0.538                 & \textbf{0.570}      \\
        Color         & 0.733                    & 0.700                 & \textbf{0.720}      \\
        Orientation   & 0.500                    & 0.567                 & \textbf{0.590}      \\
        OCR           & 0.433                    & 0.467                 & \textbf{0.490}      \\
        Size          & 0.567                    & 0.700                 & \textbf{0.730}      \\
        Position      & 0.700                    & 0.700                 & \textbf{0.720}      \\
        Counting      & 0.467                    & 0.433                 & \textbf{0.460}      \\
        \midrule
        \textbf{Average} & \textbf{0.557}         & \textbf{0.588}        & \textbf{0.613}      \\
        \bottomrule
    \end{tabular}}
\end{table}
On the VHTest YNQ benchmark, GVF Finetuning also demonstrates superior performance with an average accuracy of \textbf{0.613}, an improvement over the 0.588 achieved by the standard fine-tuned LLaVA-1.5. This further validates the method's ability to reduce hallucinations, particularly in categories like Shape (0.570 vs. 0.538), Orientation (0.590 vs. 0.567), and Size (0.730 vs. 0.700). The robust performance across both OEQ and YNQ formats underscores the comprehensive nature of GVF in improving factual adherence.

\subsubsection{Generalizability and Trade-off Analysis}
Table \ref{tab:general_benchmarks} presents the performance of the models on general multimodal benchmarks, MME and POPE.
\begin{table}[!t]\small
    \centering
    \caption{Performance on MME and POPE General Benchmarks}
    \label{tab:general_benchmarks}
    \resizebox{\linewidth}{!}{
    \begin{tabular}{lccc}
        \toprule
        Metric                     & LLaVA-1.5 & LLaVA-1.5 (Finetuned) & \textbf{Ours (GVF)} \\
        \midrule
        MME-Perception (max 2000)  & 1531.3                   & 1556.6                & \textbf{1560.1}     \\
        MME-Cognition (max 800)    & 295.4                    & 288.2                 & \textbf{287.5}      \\
        POPE (F1, mean)            & 85.9                     & 84.8                  & \textbf{85.0}       \\
        \bottomrule
    \end{tabular}}
\end{table}
Crucially, the results on MME and POPE indicate that our GVF Finetuning method not only significantly reduces visual hallucinations but also maintains, and in some cases slightly enhances, the model's general multimodal understanding and reasoning capabilities. On MME-Perception, GVF achieves \textbf{1560.1}, a slight improvement over the fine-tuned baseline's 1556.6. While MME-Cognition shows a slight decrease compared to the pre-finetuned model (287.5 vs. 295.4), it is comparable to the standard fine-tuned LLaVA-1.5 (288.2). For the POPE F1 score, GVF achieves \textbf{85.0}, which is on par with the fine-tuned baseline's 84.8 and close to the pre-finetuned LLaVA-1.5's 85.9. This demonstrates that the explicit factual guidance and loss function in GVF do not lead to a trade-off in general abilities; instead, they enable the model to become more reliable without sacrificing its broader competence.

These comprehensive results validate that GVF Finetuning successfully addresses the critical issue of visual hallucination in MLLMs by instilling a stronger adherence to objective visual facts, thereby enhancing their reliability and trustworthiness for real-world applications.

\subsection{Ablation Study of GVF Components}
To understand the individual contributions of each component within our Grounded Visual Factualization (GVF) Finetuning framework, we conducted a detailed ablation study. This involved systematically removing or modifying key elements of GVF and evaluating the resulting performance on both hallucination benchmarks and general multimodal tasks. The baseline for comparison is the \textbf{LLaVA-1.5 (Finetuned)} model, which represents standard instruction tuning without any GVF-specific mechanisms.

Our ablation variants are defined as follows:
\begin{enumerate}
    \item \textbf{LLaVA-1.5 (Finetuned)}: The standard fine-tuned model, serving as the baseline without any GVF components.
    \item \textbf{GVF w/o Factual Consistency Loss (\(\lambda=0\))}: This variant incorporates Factual Anchor Data Augmentation and Fact-Aware Instruction Tuning, but the factual consistency loss function is disabled (i.e., $\lambda=0$). This isolates the impact of explicit factual cues in the input without an explicit factual error penalty during optimization.
    \item \textbf{GVF w/o Fact-Aware Instruction Tuning}: This variant uses Factual Anchor Data Augmentation and Factual Consistency Loss, but the generated factual anchors and counter-factual prompts are not embedded into fact-aware instructions. Instead, they might be provided as supplementary information or implicitly learned, without explicit [FACT:...] or [CHECK\_COLOR:...] prefixes in the input prompt. This assesses the importance of explicit input guidance.
    \item \textbf{Ours (GVF)}: The full proposed method, integrating all three components: Factual Anchor Data Augmentation, Fact-Aware Instruction Tuning, and Factual Consistency Loss.
\end{enumerate}

Table \ref{tab:ablation_study} presents the results of our ablation study.

\begin{table*}[!t]\small
    \centering
    \caption{Ablation Study on VHTest (Average Accuracy) and General Benchmarks}
    \label{tab:ablation_study}
    % \resizebox{\linewidth}{!}{
    \begin{tabular}{lcccc}
        \toprule
        Variant                                  & VHTest OEQ Avg. Acc. & VHTest YNQ Avg. Acc. & MME-Perception & POPE F1 \\
        \midrule
        LLaVA-1.5 (Finetuned)                    & 0.296                & 0.588                & 1556.6         & 84.8    \\
        GVF w/o Fact-Aware Instruction Tuning    & 0.305                & 0.592                & 1557.0         & 84.8    \\
        GVF w/o Factual Consistency Loss (\(\lambda=0\)) & 0.320                & 0.600                & 1558.0         & 84.9    \\
        \textbf{Ours (GVF)}                      & \textbf{0.336}       & \textbf{0.613}       & \textbf{1560.1}& \textbf{85.0}   \\
        \bottomrule
    \end{tabular}
\end{table*}

The ablation study reveals several key insights:
\begin{itemize}
    \item The \textbf{Factual Anchor Data Augmentation} (implicitly present in all GVF variants) forms the foundation, as it provides the necessary structured factual signals. Even without explicit instruction tuning or FCL, just having richer, fact-grounded training data contributes to a slight improvement over the baseline (as seen by comparing LLaVA-1.5 (Finetuned) with GVF w/o Fact-Aware Instruction Tuning).
    \item \textbf{Fact-Aware Instruction Tuning} significantly enhances performance. Comparing "GVF w/o Fact-Aware Instruction Tuning" (0.305 OEQ accuracy) with "GVF w/o Factual Consistency Loss" (0.320 OEQ accuracy) shows that explicitly guiding the model's attention to factual types through structured instructions provides a substantial boost in factual accuracy. This suggests that directing the model's focus at inference time is crucial.
    \item The \textbf{Factual Consistency Loss Function} provides the final and most impactful boost in mitigating hallucinations. The full GVF model (0.336 OEQ accuracy) outperforms "GVF w/o Factual Consistency Loss" (0.320 OEQ accuracy) by a notable margin. This demonstrates that explicitly penalizing factual errors during training is essential for instilling robust factual adherence and minimizing the generation of erroneous details.
    \item Importantly, all GVF variants maintain or slightly improve performance on general benchmarks (MME and POPE), indicating that these factual grounding mechanisms do not come at the cost of general multimodal understanding.
\end{itemize}
This systematic analysis confirms that all three core components of GVF Finetuning synergistically contribute to its superior performance in reducing visual hallucinations.

\subsection{Impact of Factual Consistency Loss Weight}
The hyperparameter $\lambda$ in Equation \ref{eq:total_loss} plays a critical role in balancing the influence of the standard cross-entropy loss ($L_{CE}$) and our proposed factual consistency loss ($L_{FCL}$). We investigated the sensitivity of GVF's performance to different values of $\lambda$ to identify an optimal balance that maximizes factual consistency without compromising general model capabilities.

\begin{figure}[!t]
    \centering
    \includegraphics[width=\linewidth]{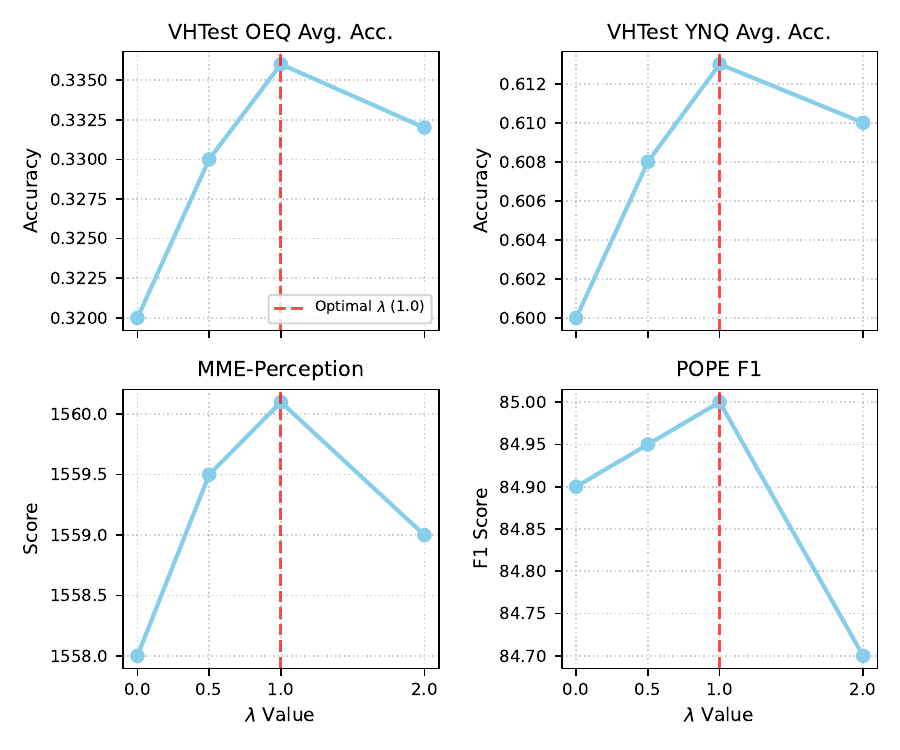} 
    \caption{Sensitivity Analysis of Factual Consistency Loss Weight (\(\lambda\))}
    \label{fig:lambda_sensitivity} 
\end{figure}

Our experiments varied $\lambda$ across a range of values, from 0 (effectively disabling $L_{FCL}$) to higher values that increase its penalty weight. The results are summarized in Figure \ref{fig:lambda_sensitivity}.

As observed in Figure \ref{fig:lambda_sensitivity}, setting $\lambda=0.0$ (which corresponds to "GVF w/o Factual Consistency Loss" from the ablation study) yields a solid performance, benefiting from factual data augmentation and fact-aware instructions. However, as $\lambda$ increases, the model's factual consistency, particularly on VHTest OEQ and YNQ, steadily improves, reaching its peak at $\lambda=1.0$. This indicates that a moderate and balanced penalty for factual inconsistencies is highly effective in guiding the model towards more accurate responses.

Beyond $\lambda=1.0$, for instance at $\lambda=2.0$, we observe a slight decrease in performance across most metrics. While VHTest accuracy remains competitive, there is a marginal drop in MME-Perception and POPE F1 scores. This suggests that an excessively high $\lambda$ can potentially lead to an over-emphasis on strict factual adherence, possibly making the model overly cautious or impacting its ability to generate nuanced or diverse responses that are still factually plausible. An overly aggressive penalty might also hinder the learning of broader semantic understanding by prioritizing specific factual correctness above all else. Based on these findings, $\lambda=1.0$ was selected as the optimal weight for the factual consistency loss in our full GVF model, providing the best balance between hallucination mitigation and general performance.

\subsection{Qualitative Analysis and Error Patterns}
\begin{figure}[!t]
    \centering
    \includegraphics[width=\linewidth]{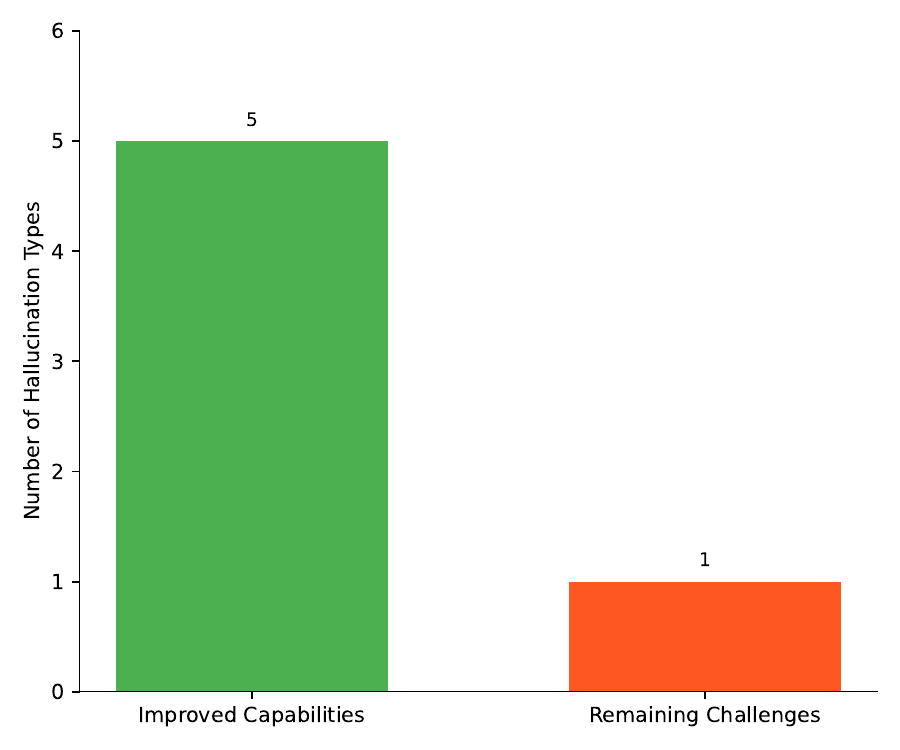}
    \caption{Qualitative Comparison and Error Patterns}
    \label{fig:qualitative_analysis}
\end{figure}

To complement the quantitative results, we conducted a qualitative analysis of model outputs from the baselines and our GVF Finetuning method. This allowed us to gain deeper insights into the types of visual hallucinations mitigated and the remaining challenges. Figure \ref{fig:qualitative_analysis} illustrates characteristic improvements observed with GVF and highlights common error patterns.

Our qualitative analysis confirms that GVF Finetuning significantly reduces various types of visual hallucinations, particularly for objective facts such as counting, color, existence, position, and OCR. The model demonstrates a heightened awareness of visual details and a reduced tendency to fabricate information. For instance, in counting tasks, GVF is less prone to off-by-one errors or gross overestimations. For existence questions, it confidently rejects the presence of non-existent objects, a common hallucination in ungrounded MLLMs.

Despite these substantial improvements, certain challenges remain. While GVF excels at grounding objective facts, more complex reasoning tasks that involve nuanced interactions between multiple objects, inferring intent, or understanding abstract concepts based on visual cues can still present difficulties. The model might provide factually correct descriptions of individual elements but could still struggle with the holistic interpretation of a complex scene or the subtle implications of visual information. This suggests that future work could focus on extending factual grounding to higher-level conceptual and relational facts, building upon the strong foundation of objective factual consistency established by GVF.

\section{Conclusion}
This paper addressed the critical challenge of visual hallucination (VH) in Multimodal Large Language Models (MLLMs), a pervasive issue compromising their reliability. We proposed \textbf{Grounded Visual Factualization (GVF) Finetuning}, a novel and systematic approach designed to instill deeper adherence to objective visual facts through three synergistic components: Factual Anchor Data Augmentation, Fact-Aware Instruction Tuning, and a Factual Consistency Loss function. Our extensive experiments, utilizing LLaVA-1.5-13B, quantitatively demonstrated GVF's superior efficacy, achieving significant performance improvements on the challenging VHTest benchmark across all eight visual hallucination types. Specifically, GVF Finetuning elevated the average Open-Ended Question accuracy from 0.296 to a notable \textbf{0.336}, with strong gains in categories like Shape, Position, Counting, and OCR, while crucially maintaining or enhancing performance on general multimodal benchmarks. Ablation studies further elucidated the paramount contribution of the Factual Consistency Loss. 
    
{\small
\bibliographystyle{unsrtnat}
\bibliography{main}}

\begin{thebibliography}{28}
\providecommand{\natexlab}[1]{#1}
\providecommand{\url}[1]{\texttt{#1}}
\expandafter\ifx\csname urlstyle\endcsname\relax
  \providecommand{\doi}[1]{doi: #1}\else
  \providecommand{\doi}{doi: \begingroup \urlstyle{rm}\Url}\fi

\bibitem[Masry et~al.(2022)Masry, Long, Tan, Joty, and Hoque]{masry2022chartq}
Ahmed Masry, Do~Xuan Long, Jia~Qing Tan, Shafiq Joty, and Enamul Hoque.
\newblock {{C}}hart{{QA}}: A benchmark for question answering about charts with visual and logical reasoning.
\newblock In \emph{Findings of the Association for Computational Linguistics: ACL 2022}, pages 2263--2279. Association for Computational Linguistics, 2022.
\newblock \doi{10.18653/v1/2022.findings-acl.177}.

\bibitem[Oguz et~al.(2022)Oguz, Chen, Karpukhin, Peshterliev, Okhonko, Schlichtkrull, Gupta, Mehdad, and Yih]{oguz2022unikqa}
Barlas Oguz, Xilun Chen, Vladimir Karpukhin, Stan Peshterliev, Dmytro Okhonko, Michael Schlichtkrull, Sonal Gupta, Yashar Mehdad, and Scott Yih.
\newblock {{U}}ni{{K}}-{{QA}}: Unified representations of structured and unstructured knowledge for open-domain question answering.
\newblock In \emph{Findings of the Association for Computational Linguistics: NAACL 2022}, pages 1535--1546. Association for Computational Linguistics, 2022.
\newblock \doi{10.18653/v1/2022.findings-naacl.115}.

\bibitem[McKenna et~al.(2023)McKenna, Li, Cheng, Hosseini, Johnson, and Steedman]{mckenna2023source}
Nick McKenna, Tianyi Li, Liang Cheng, Mohammad Hosseini, Mark Johnson, and Mark Steedman.
\newblock Sources of hallucination by large language models on inference tasks.
\newblock In \emph{Findings of the Association for Computational Linguistics: EMNLP 2023}, pages 2758--2774. Association for Computational Linguistics, 2023.
\newblock \doi{10.18653/v1/2023.findings-emnlp.182}.

\bibitem[Ji et~al.(2023)Ji, Yu, Xu, Lee, Ishii, and Fung]{ji2023toward}
Ziwei Ji, Tiezheng Yu, Yan Xu, Nayeon Lee, Etsuko Ishii, and Pascale Fung.
\newblock Towards mitigating {{LLM}} hallucination via self reflection.
\newblock In \emph{Findings of the Association for Computational Linguistics: EMNLP 2023}, pages 1827--1843. Association for Computational Linguistics, 2023.
\newblock \doi{10.18653/v1/2023.findings-emnlp.123}.

\bibitem[Hu et~al.(2021)Hu, Liu, Zhao, and Jin]{hu2021mmgcn}
Jingwen Hu, Yuchen Liu, Jinming Zhao, and Qin Jin.
\newblock {{MMGCN}}: Multimodal fusion via deep graph convolution network for emotion recognition in conversation.
\newblock In \emph{Proceedings of the 59th Annual Meeting of the Association for Computational Linguistics and the 11th International Joint Conference on Natural Language Processing (Volume 1: Long Papers)}, pages 5666--5675. Association for Computational Linguistics, 2021.
\newblock \doi{10.18653/v1/2021.acl-long.440}.

\bibitem[Chen et~al.(2021)Chen, Tang, Qin, Liang, Liu, Xing, and Lin]{chen2021geoqa}
Jiaqi Chen, Jianheng Tang, Jinghui Qin, Xiaodan Liang, Lingbo Liu, Eric Xing, and Liang Lin.
\newblock {{G}}eo{{QA}}: A geometric question answering benchmark towards multimodal numerical reasoning.
\newblock In \emph{Findings of the Association for Computational Linguistics: ACL-IJCNLP 2021}, pages 513--523. Association for Computational Linguistics, 2021.
\newblock \doi{10.18653/v1/2021.findings-acl.46}.

\bibitem[Lin et~al.(2024{\natexlab{a}})Lin, Zhang, Tian, Yu, and Lan]{lin2024dpl}
Zhihao Lin, Qi~Zhang, Zhen Tian, Peizhuo Yu, and Jianglin Lan.
\newblock Dpl-slam: enhancing dynamic point-line slam through dense semantic methods.
\newblock \emph{IEEE Sensors Journal}, 24\penalty0 (9):\penalty0 14596--14607, 2024{\natexlab{a}}.

\bibitem[Lin et~al.(2024{\natexlab{b}})Lin, Tian, Zhang, Zhuang, and Lan]{lin2024enhanced}
Zhihao Lin, Zhen Tian, Qi~Zhang, Hanyang Zhuang, and Jianglin Lan.
\newblock Enhanced visual slam for collision-free driving with lightweight autonomous cars.
\newblock \emph{Sensors}, 24\penalty0 (19):\penalty0 6258, 2024{\natexlab{b}}.

\bibitem[Song et~al.(2022)Song, Dong, Zhang, Liu, and Wei]{song2022clip}
Haoyu Song, Li~Dong, Weinan Zhang, Ting Liu, and Furu Wei.
\newblock {{CLIP}} models are few-shot learners: Empirical studies on {{VQA}} and visual entailment.
\newblock In \emph{Proceedings of the 60th Annual Meeting of the Association for Computational Linguistics (Volume 1: Long Papers)}, pages 6088--6100. Association for Computational Linguistics, 2022.
\newblock \doi{10.18653/v1/2022.acl-long.421}.

\bibitem[Zhang et~al.(2024{\natexlab{a}})Zhang, Chen, Hua, and Luo]{zhang2024fate}
Fan Zhang, Chong Chen, Xian-Sheng Hua, and Xiao Luo.
\newblock Fate: Learning effective binary descriptors with group fairness.
\newblock \emph{IEEE Transactions on Image Processing}, 33:\penalty0 3648--3661, 2024{\natexlab{a}}.

\bibitem[Zhao et~al.(2021)Zhao, Lu, and Lee]{zhao2021sparta}
Tiancheng Zhao, Xiaopeng Lu, and Kyusong Lee.
\newblock {{SPARTA}}: Efficient open-domain question answering via sparse transformer matching retrieval.
\newblock In \emph{Proceedings of the 2021 Conference of the North American Chapter of the Association for Computational Linguistics: Human Language Technologies}, pages 565--575. Association for Computational Linguistics, 2021.
\newblock \doi{10.18653/v1/2021.naacl-main.47}.

\bibitem[Hendricks and Nematzadeh(2021)]{hendricks2021probin}
Lisa~Anne Hendricks and Aida Nematzadeh.
\newblock Probing image-language transformers for verb understanding.
\newblock In \emph{Findings of the Association for Computational Linguistics: ACL-IJCNLP 2021}, pages 3635--3644. Association for Computational Linguistics, 2021.
\newblock \doi{10.18653/v1/2021.findings-acl.318}.

\bibitem[Zhou et~al.(2023)Zhou, Geng, Shen, Tao, Long, Lou, and Shen]{zhou2023thread}
Yucheng Zhou, Xiubo Geng, Tao Shen, Chongyang Tao, Guodong Long, Jian-Guang Lou, and Jianbing Shen.
\newblock Thread of thought unraveling chaotic contexts.
\newblock \emph{arXiv preprint arXiv:2311.08734}, 2023.

\bibitem[Zhou et~al.(2025{\natexlab{a}})Zhou, Shen, and Cheng]{zhou2025weak}
Yucheng Zhou, Jianbing Shen, and Yu~Cheng.
\newblock Weak to strong generalization for large language models with multi-capabilities.
\newblock In \emph{The Thirteenth International Conference on Learning Representations}, 2025{\natexlab{a}}.

\bibitem[Zhang et~al.(2024{\natexlab{b}})Zhang, Hua, Chen, and Luo]{zhang2024statistical}
Fan Zhang, Xian-Sheng Hua, Chong Chen, and Xiao Luo.
\newblock A statistical perspective for efficient image-text matching.
\newblock In \emph{Proceedings of the 2024 Conference of the North American Chapter of the Association for Computational Linguistics: Human Language Technologies (Volume 1: Long Papers)}, pages 355--369, 2024{\natexlab{b}}.

\bibitem[Deng et~al.(2021)Deng, Awadallah, Meek, Polozov, Sun, and Richardson]{deng2021struct}
Xiang Deng, Ahmed~Hassan Awadallah, Christopher Meek, Oleksandr Polozov, Huan Sun, and Matthew Richardson.
\newblock Structure-grounded pretraining for text-to-{{SQL}}.
\newblock In \emph{Proceedings of the 2021 Conference of the North American Chapter of the Association for Computational Linguistics: Human Language Technologies}, pages 1337--1350. Association for Computational Linguistics, 2021.
\newblock \doi{10.18653/v1/2021.naacl-main.105}.

\bibitem[Nan et~al.(2021)Nan, Nogueira~dos Santos, Zhu, Ng, McKeown, Nallapati, Zhang, Wang, Arnold, and Xiang]{nan2021improv}
Feng Nan, Cicero Nogueira~dos Santos, Henghui Zhu, Patrick Ng, Kathleen McKeown, Ramesh Nallapati, Dejiao Zhang, Zhiguo Wang, Andrew~O. Arnold, and Bing Xiang.
\newblock Improving factual consistency of abstractive summarization via question answering.
\newblock In \emph{Proceedings of the 59th Annual Meeting of the Association for Computational Linguistics and the 11th International Joint Conference on Natural Language Processing (Volume 1: Long Papers)}, pages 6881--6894. Association for Computational Linguistics, 2021.
\newblock \doi{10.18653/v1/2021.acl-long.536}.

\bibitem[Honovich et~al.(2021)Honovich, Choshen, Aharoni, Neeman, Szpektor, and Abend]{honovich2021q2}
Or~Honovich, Leshem Choshen, Roee Aharoni, Ella Neeman, Idan Szpektor, and Omri Abend.
\newblock $q^{{2}}$: {{E}}valuating factual consistency in knowledge-grounded dialogues via question generation and question answering.
\newblock In \emph{Proceedings of the 2021 Conference on Empirical Methods in Natural Language Processing}, pages 7856--7870. Association for Computational Linguistics, 2021.
\newblock \doi{10.18653/v1/2021.emnlp-main.619}.

\bibitem[Schuster et~al.(2021)Schuster, Fisch, and Barzilay]{schuster2021get}
Tal Schuster, Adam Fisch, and Regina Barzilay.
\newblock Get your vitamin {{C}}! robust fact verification with contrastive evidence.
\newblock In \emph{Proceedings of the 2021 Conference of the North American Chapter of the Association for Computational Linguistics: Human Language Technologies}, pages 624--643. Association for Computational Linguistics, 2021.
\newblock \doi{10.18653/v1/2021.naacl-main.52}.

\bibitem[Ovadia et~al.(2024)Ovadia, Brief, Mishaeli, and Elisha]{ovadia2024finetu}
Oded Ovadia, Menachem Brief, Moshik Mishaeli, and Oren Elisha.
\newblock Fine-tuning or retrieval? comparing knowledge injection in {{LLM}}s.
\newblock In \emph{Proceedings of the 2024 Conference on Empirical Methods in Natural Language Processing}, pages 237--250. Association for Computational Linguistics, 2024.
\newblock \doi{10.18653/v1/2024.emnlp-main.15}.

\bibitem[Zhou et~al.(2025{\natexlab{b}})Zhou, Song, and Shen]{zhou2025improving}
Yucheng Zhou, Lingran Song, and Jianbing Shen.
\newblock Improving medical large vision-language models with abnormal-aware feedback.
\newblock \emph{arXiv preprint arXiv:2501.01377}, 2025{\natexlab{b}}.

\bibitem[Wan and Bansal(2022)]{wan2022factpe}
David Wan and Mohit Bansal.
\newblock {{F}}act{{PEGASUS}}: Factuality-aware pre-training and fine-tuning for abstractive summarization.
\newblock In \emph{Proceedings of the 2022 Conference of the North American Chapter of the Association for Computational Linguistics: Human Language Technologies}, pages 1010--1028. Association for Computational Linguistics, 2022.
\newblock \doi{10.18653/v1/2022.naacl-main.74}.

\bibitem[Liu et~al.(2022)Liu, Ji, Fu, Tam, Du, Yang, and Tang]{liu2022ptunin}
Xiao Liu, Kaixuan Ji, Yicheng Fu, Weng Tam, Zhengxiao Du, Zhilin Yang, and Jie Tang.
\newblock {{P}}-tuning: Prompt tuning can be comparable to fine-tuning across scales and tasks.
\newblock In \emph{Proceedings of the 60th Annual Meeting of the Association for Computational Linguistics (Volume 2: Short Papers)}, pages 61--68. Association for Computational Linguistics, 2022.
\newblock \doi{10.18653/v1/2022.acl-short.8}.

\bibitem[Zhou et~al.(2024)Zhou, Li, Wang, and Shen]{zhou2024visual}
Yucheng Zhou, Xiang Li, Qianning Wang, and Jianbing Shen.
\newblock Visual in-context learning for large vision-language models.
\newblock In \emph{Findings of the Association for Computational Linguistics, {ACL} 2024, Bangkok, Thailand and virtual meeting, August 11-16, 2024}, pages 15890--15902. Association for Computational Linguistics, 2024.

\bibitem[Zhang et~al.(2025)Zhang, Wang, Cheng, Peng, Wang, Xiao, Chen, Hua, and Luo]{zhang2025dream}
Fan Zhang, Changhu Wang, Zebang Cheng, Xiaojiang Peng, Dongjie Wang, Yijia Xiao, Chong Chen, Xian-Sheng Hua, and Xiao Luo.
\newblock Dream: Decoupled discriminative learning with bigraph-aware alignment for semi-supervised 2d-3d cross-modal retrieval.
\newblock In \emph{Proceedings of the AAAI Conference on Artificial Intelligence}, volume~39, pages 13206--13214, 2025.

\bibitem[Wang et~al.(2024)Wang, Zhu, and Liang]{wang2024virtual}
Peng Wang, ZQ~Zhu, and Dawei Liang.
\newblock Virtual back-emf injection based online parameter identification of surface-mounted pmsms under sensorless control.
\newblock \emph{IEEE Transactions on Industrial Electronics}, 2024.

\bibitem[Wang et~al.(2025)Wang, Zhu, and Feng]{wang2025virtual}
Peng Wang, ZQ~Zhu, and Zhibin Feng.
\newblock Virtual back-emf injection-based online full-parameter estimation of dtp-spmsms under sensorless control.
\newblock \emph{IEEE Transactions on Transportation Electrification}, 2025.

\bibitem[Zhang et~al.(2022)Zhang, Strubell, and Hovy]{zhang2022a}
Zhisong Zhang, Emma Strubell, and Eduard Hovy.
\newblock A survey of active learning for natural language processing.
\newblock In \emph{Proceedings of the 2022 Conference on Empirical Methods in Natural Language Processing}, pages 6166--6190. Association for Computational Linguistics, 2022.
\newblock \doi{10.18653/v1/2022.emnlp-main.414}.

\end{thebibliography}


\end{document}